\documentclass[12pt, a4paper]{article}
\usepackage{amsmath,amsfonts,amsthm,mathrsfs,bm}
\usepackage[mathlines]{lineno}

\numberwithin{equation}{section}
\newtheorem{theorem}{Theorem}
\newtheorem{lemma}{Lemma}

\newtheorem{corollary}{Corollary}
\newtheorem{definition}{Definition}

\usepackage[scale={0.72,0.743}, hmarginratio=1:1, vmarginratio=1:1, headheight=2ex, headsep = 2ex, footskip = 3.9ex, nomarginpar]{geometry}

\def\NN{\mathbb N}
\def\ZZ{\mathbb Z}
\def\RR{\mathbb R}

\def\EE{\mathbb E}
\def\TT{\mathbb T}

\begin{document}
\title{Rates of Approximation by ReLU Shallow Neural Networks}
\author{Tong Mao \\
Institute of Mathematical Sciences, Claremont Graduate University \\
710 N. College Avenue, Claremont, CA 91711, USA \\
Email: tong.mao@cgu.edu  \\
Ding-Xuan Zhou \\
School of Mathematics and Statistics, University of Sydney \\
Sydney, NSW 2006, Australia \\
Email: dingxuan.zhou@sydney.edu.au}
\date{}

\maketitle
\begin{abstract}
Neural networks activated by the rectified linear unit (ReLU) play a central role in the recent development of deep learning. The topic of approximating functions from H\"older spaces by these networks is crucial for understanding the efficiency of the induced learning algorithms. Although the topic has been well investigated in the setting of deep neural networks with many layers of hidden neurons, it is still open for shallow networks having only one hidden layer. 
In this paper, we provide rates of uniform approximation by these networks. 
We show that ReLU shallow neural networks  with $m$ hidden neurons can uniformly approximate functions from the H\"older space $W_\infty^r([-1, 1]^d)$ with rates $O((\log m)^{\frac{1}{2} +d}m^{-\frac{r}{d}\frac{d+2}{d+4}})$ when $r<d/2 +2$. Such rates are very close to the optimal one $O(m^{-\frac{r}{d}})$ in the sense that $\frac{d+2}{d+4}$ is close to $1$, when the dimension $d$ is large.
\end{abstract}

\noindent {\it Keywords}: deep learning, shallow neural networks, ReLU, rates of uniform approximation, H\"older space

\noindent {\it Mathematics Subject Classification (2020)}: 68T07, 41A25, 68Q32

\baselineskip 16pt

\section{Introduction} 

The exploration of approximating functions by neural networks has a history of over 30 years. 
Let $x=(x_1, \dots, x_d)\in\RR^d$ be the data vector and $m\in\NN$. A shallow neural network of width $m$ associated with a continuous activation function $\sigma: \RR \to \RR$ is defined by
\begin{equation}\label{defshallow}
f_m(x)=\sum\limits_{i=1}^m\beta_i\sigma(\alpha_i\cdot x-b_i),
\end{equation}
where $\{\alpha_i\}_{i=1}^m\subset\RR^d$ are connection vectors, $\{\beta_i\}_{i=1}^m\subset\RR$ weights and $\{b_i\}_{i=1}^m\subset\RR$ biases. 
The universality of shallow networks \cite{Cybenko,Hornik,Pinkus} asserts for any non-polynomial activation that any continuous function on any compact subset of $\RR^d$ can be approximated by output functions of the form (\ref{defshallow}) to an arbitrary accuracy when the number $m$ of hidden neurons is large enough. 
Rates of approximation by such output functions from the hypothesis space 
\begin{equation}\label{defspace}
H_m=\left\{\sum\limits_{i=1}^m\beta_i\sigma(\alpha_i\cdot x-b_i):\ \alpha_i\in\RR^d,\ \beta_i\in\RR,\ b_i\in\RR\right\}.
\end{equation}
were also studied in a large literature when $\sigma$ is a sigmoid type $C^\infty$ activation function. In \cite{Barron1993} it was proved that a function $f$ on $\RR^d$ with its Fourier transform $\hat f$ satisfying $\int_{\RR^d}\lvert\omega\rvert\lvert\hat f(\omega)\rvert d\omega<\infty$ can be approximated uniformly on $[-1, 1]^d$ by $f_m\in H_m$ with rates $O(m^{-1/2})$. The error rate is proved  (e.g. \cite{Mhaskar}) to be $O(m^{-r/d})$ for functions from the H\"older space $W_\infty^r([-1, 1]^d)$ defined as follows.
\begin{definition}
Let $r\in\NN$, $d\in\NN$, and $\Omega\subset\RR^d$ be a compact domain with non-empty interior. The H\"older space $W_\infty^r(\Omega)$ is defined by
\begin{equation}\label{defSobolev}
W_\infty^r(\Omega)=\{f \in C(\Omega):\ \max\limits_{0\leq\lVert\alpha\rVert_1\leq r}\lVert D^\alpha f\rVert_{L_\infty(\Omega)}<\infty\}
\end{equation}
with the norm $\lVert f\rVert_{W_\infty^r(\Omega)}=\max\limits_{0\leq\lVert\alpha\rVert_1\leq r}\lVert D^\alpha f\rVert_{L_\infty(\Omega)}$ defined with the partial derivatives $D^\alpha f$ of order $\alpha=(\alpha_1, \cdots, \alpha_d) \in \ZZ_+^d$.
\end{definition}

 When training network parameters with gradient descent methods in deep learning, the classical neural networks with sigmoid type $C^\infty$ activation functions often encounter the {\bf gradient vanishing problem}. To solve this problem, in most neural networks used for deep learning applications, the classical sigmoid type $C^\infty$ activation functions are replaced by the \textbf{rectified linear unit} (ReLU) $\sigma$ defined as 
$$\sigma(x)=\max\{0,x\},\qquad x\in\RR.$$

Within the last few years, rates of function approximation by ReLU deep neural networks have been obtained when the network has many layers with depth increasing with the number $m$ of hidden neurons. D. Yarotsky \cite{Yarotsky} observed that iterates of ReLU can be used to realize piecewise linear interpolations of the univariate quadratic function $\varphi(u) = u^2$ with few parameters. Based on this observation, he proved that ReLU deep neural networks with $O(\log m)$ layers and $O(m\log m)$ neurons can achieve an almost optimal error rate $O(m^{-r/d})$ for approximating functions from $W_\infty^r([-1, 1]^d)$. For approximating smooth functions and piecewise-smooth functions, upper bounds were given in \cite{PetersenVoigtlaender, Grohs, Guehring, HanYuLinZhou}. Applying Yarotsky's method to functions from the Korobov space, a deep neural network with $O(\log m)$ layers and $O(m(\log m)^{\frac{3}{2}(d-1)+1})$ neurons, which achieves the rate $O(m^{-2})$, was constructed in \cite{MontanelliDu}. 

All the above mentioned results are for ReLU deep neural networks with many layers. Furthermore, in \cite{ShahamCloningerCoifman}, it was proved that on a smooth $d$-dimensional compact manifold without boundary, $C^2$ functions can be approximated by a deep network of depth $4$ with a rate $O(m^{-2/d})$, which is optimal. Rates of approximation by networks induced by ReLU type activation functions such as $\left(\max\{0,x\}\right)^\alpha$ with $\alpha >1$ were also obtained in \cite{Bach2017, Mhaskar2019}. This recent literaure on ReLU type networks and  
the classical work on sigmoid shallow networks lead to a  natural open question whether one can derive rates of function approximation by ReLU shallow networks of one hidden layer.

The purpose of this paper is to answer the above open question and present rates of approximating functions from H\"older spaces uniformly by \textbf{ReLU shallow neural networks}. An approximation theory about shallow neural networks plays a fundamental role in understanding ReLU networks as a benchmark. It has some other applications such as those for convolutional neural networks to be discussed below. Similar analysis was conducted in \cite[Theorem 4.1]{MhaskarPoggio} for sigmoid networks which demonstrates how to deduce approximation rates of a sigmoid deep neural network by using those of sigmoid shallow networks. 

A nice study for ReLU shallow networks was carried out by Klusowski and Barron \cite{KlusowskiBarron}. They gave rates $O(m^{-\frac{1}{2}-\frac{1}{d}})$ of approximating functions $f$ whose Fourier transform $\hat{f}$ satisfies $\int_{\RR^d}\lvert\hat f(\omega)\rvert\lVert\omega\rVert_1^2d\omega<\infty$. If we want to apply this result to the H\"older space $W_\infty^r(\Omega)$, then the regularity index $r$ must satisfy $r>d/2$, as discussed in \cite{Zhou}, which might be too large for some learning problems dealing with data of large dimensions $d >>1$.

A possible way to overcome this barrier is using an intermediate function $f_R(x)=\int_{\lvert\omega\rvert\leq R}\hat f(\omega)e^{i\omega\cdot x}d\omega$ and estimate the error $\lVert f-f_R\rVert_\infty$ and $\lVert f_R-f_m\rVert_\infty$, where $f_m$ is the function in \cite{KlusowskiBarron} which approximates $f_R$. However, this method does not make full use of frequency domain properties of the Fourier transform. One can only obtain a rate $O(m^{-\frac{r}{2d}})$, which is much worse than the optimal rate $O(m^{-r/d})$.

In this paper, we use a novel idea motivated by tools from Fourier analysis \cite{Fefferman,Trigub, Carleson} to carry out time-frequency analysis for partial sums of Fourier series and decompose the error into multi-level parts according to various frequecy levels. Then we are able to show that the error of approximating functions from $W_\infty^r([-1,1]^d)$ by ReLU shallow neural networks can be estimated with rates $O((\log m)^{\frac{1}{2}+d}m^{-\frac{r}{d}\frac{d+2}{d+4}})$ when $r<d/2+2$. This rate of approximation is very close to the optimal one $O(m^{-r/d})$ when the dimension $d$ is large. In fact, $O(m^{-r/d})$ is the lower bound for the approximation by any neural network with $m$ parameters, which was proved in \cite{DeVoreHowardMicchelli} and will be discussed in Section \ref{discussion}. Throughout the paper we take the domain $D=[-1, 1]^d$ of functions for approximation and $\sigma$ to be the ReLU activation function.

\section{Main Results}\label{Mainresults}
The following theorem, to be proved in Section \ref{proof of main results}, is our first main result. It gives an upper bound for approximating functions from $W_\infty^r(D)$ uniformly by ReLU shallow neural networks.
\begin{theorem}\label{mainresult}
Let $d, r\in\NN$. Then there exists a constant $C(d,r)$ depending only on $d$ and $r$ such that for any $F\in W^r_\infty(D)$ and $m\in\mathbb N$, there holds
\begin{equation}\label{maininequation}
\begin{split}
\inf\limits_{f_m\in H_m}\lVert F-f_m\rVert_{L_\infty(D)}\leq C(d,r)\lVert F\rVert_{W_\infty^r(D)}\left\{\begin{array}{ll}
(\log m)^{\frac{1}{2}+d}m^{-\frac{r}{d}\frac{d+2}{d+4}},\qquad&\hbox{if }r<\frac{d}{2}+2,\\
(\log m)^{\frac{3}{2}+d}m^{-\frac{r}{d}\frac{d+2}{d+4}},\qquad&\hbox{if }r=\frac{d}{2}+2,\\
(\log m)^{\frac{1}{2}}m^{-\frac{1}{2}-\frac{1}{d}},\qquad&\hbox{if }r>\frac{d}{2}+2.\end{array}\right.
\end{split}
\end{equation}
\end{theorem}

As all the existing approximation results on ReLU networks are either for deep networks with multiple layers or for shallow networks \cite{KlusowskiBarron} with a large regularity index $r>d/2$, Theorem \ref{mainresult} establishes the first approximation theory with almost optimal rates for approximating functions from H$\ddot{\mathrm{o}}$lder spaces by ReLU shallow networks.

Theorem \ref{mainresult} holds true only for the H\"older space $W_\infty^r(D)$ with an integer regularity index $r$. The restriction on $r$ is due to a technique in our proof for estimating a quantity involving multplications of a Fourier series with polynomial sequences. It would be interesting to extend Theorem \ref{mainresult} to H\"older spaces with non-integer regularity indices and Sobolev spaces $W_p^r(D)$ which would allow H\"older and Sobolev spaces to be defined spectrally \cite{Chandrasekaran} and thereby more general approximation estimates given in terms of moduli of smoothness. 

Theorem \ref{mainresult} may be applied to various ReLU networks. One example is the important family of convolutional neural networks (CNNs), which are widely appiled in speech recognition, image classification and many other tasks \cite{HintonOsinderoTeh,KrizhevskySutskeverHinton}. 

Given a sequence $w=(w_k)_{k\in\ZZ}$ on $\ZZ$ supported in
$\{0, 1, \ldots, s\}$ and another one $x=(x_k)_{k\in\ZZ}$ supported in $\{1, 2, \ldots, t\}$, the convolution of $w$ and $x$ is a sequence supported in $\{1,2,\dots,t+s\}$ given by
$$ \left(w{*} x\right)_i = \sum_{k\in\ZZ} w_{i-k} x_k = \sum_{k=1}^t w_{i-k} x_k, \qquad i\in\ZZ. $$
This convolutional operation induces a convolutional Toeplitz matrix
$$T^w:=(w_{i-k})_{1\leq i\leq t+s,1\leq k\leq t}$$
with the size $(t+s)\times t$ depending on that of the input vector in $\RR^t$ which corresponds to the zero-padding approach in convolutional networks.

\begin{definition}\label{defDCNN}
Let $x=(x_1, \ldots, x_d)\in\RR^d$ be the input data vector, $s, J \in\NN$, and $\{d_j\}_{j=1}^{J}$ given by $d_0 =d$, 
$$ d_j = d_{j-1} + s, \qquad j\in\{1, \ldots, J\}.$$
The \textbf{deep CNN} $\{h^{(j)}: \RR^d\to \RR^{d_j}\}_{j=1}^{J}$ with widths $\{d_j\}_{j=1}^{J}$, sequences of filters ${\bf w} :=\{w^{(j)}:\ \ZZ\to\RR\}_{j=1}^{J}$ supported in $\{0,\dots,s\}$, and biases ${\bf b}:=\{b^{(j)}\in\RR^{d_j}\}_{j=1}^{J}$ is defined by
\begin{equation}
h^{(j)}(x)=\mathcal{A}_{j}\circ \ldots \circ \mathcal{A}_{1}(x),\qquad j=1, \ldots, J,
\end{equation}
where $\mathcal{A}_{j}$ is a map from $\RR^{d_{j-1}}$ to $\RR^{d_j}$ defined 
with the $d_j\times d_{j-1}$ convolutional matrix $T^{w^{(j)}}$ by acting $\sigma$ componentwise as
$$\mathcal{A}_{j}(v)=\sigma(T^{w^{(j)}}v-b^{(j)}),\qquad v\in\mathbb{R}^{d_{j-1}}.$$
\end{definition}
It was proved in \cite{Zhou} that when $r>d/2+2$, deep CNNs can approximate functions from $W_\infty^r(D)$ uniformly with rate $O(\sqrt{\log J}J^{-1/2-1/d})$. In this paper, we can use Theorem 1 to derive rates of approximating by deep CNNs functions from the H\"older space with a smaller index $r\leq d/2+2$, which is our second main result, to be proved in Section \ref{proof of main results}. 

\begin{corollary}\label{dcnncorollary}
Let $d,\ s\in\NN$ and $F\in W_\infty^r(D)$ for some integer $r\in\NN$. Then for any $J\in\NN$, there exist $\mathbf{w}=\{w^{(j)}\}_{j=1}^J$ and $\mathbf{b}=\{b^{(j)}\}_{j=1}^J$ such that the space 
$$\mathcal H_J^{\mathbf{w,b}}=\left\{c\cdot h^{(J)}(x):\ c\in\RR^{d_J}\right\}$$
contains an element $f_J^{\mathbf{w,b}}$ satisfying
$$\lVert F-f_J^{\mathbf{w,b}}\rVert_{C(D)}\leq\left\{\begin{array}{ll}
 C_1(d,r)\lVert F\rVert_{W_\infty^r(D)}(\log J)^{d+1/2}J^{-\frac{r}{d}\frac{d+2}{d+4}},\qquad&\hbox{if }r<\frac{d}{2}+2,\\
 C_1(d,r)\lVert F\rVert_{W_\infty^r(D)}(\log J)^{d+3/2}J^{-\frac{r}{d}\frac{d+2}{d+4}},\qquad&\hbox{if }r=\frac{d}{2}+2,\\
 C_1(d,r)\lVert F\rVert_{W_\infty^r(D)}(\log J)^{1/2}J^{-\frac{1}{2}-\frac{1}{d}},\qquad&\hbox{if }r>\frac{d}{2}+2,
\end{array}\right.$$
where $C_1(d,r)$ depends only on $d$ and $r$.
\end{corollary}

The rates of approximation presented in Corollary \ref{dcnncorollary} are stated in terms of the depth $J$ or number of layers of the CNN. In many applications of CNNs, the depth $J$ is large. Some relations between deep CNNs of large depth and fully-connected networks have been observed recently. In \cite{ZhouDownsampling}, it was proved that the last layer of any fully-connected network is identical to that of a deep CNN with at most $8$ times number of free parameters. For approximating or learning ridge function \cite{FangFengHuangZhou}, radial functions \cite{MaoShiZhou}, and functions from Korobov spaces \cite{MaoZhou}, deep CNNs can be achieve the same accuracy with much smaller number of free parameters than fully-connected networks. In a recent application of CNNs to readability of Chinese texts \cite{FengHWZ}, it is found that one layer or two is already efficient. Conducting analysis for approximation and learning by CNNs with small depths would be an interesting task.

\section{Error Decomposition and Preliminary Analysis}
To prove Theorem \ref{mainresult}, we need an error decomposition: first we extend $F$ to a $2\pi$-periodic function $f$ on $\RR^d$, then we decompose the error between $F$ and $f_m$ into two parts, involving the (high-order) Jackson operator. 

By the well-known extension theorem (e.g. \cite[Chapter 6]{Stein}), $F$ can be extended to a $2\pi$-periodic continuous function $f$ on $\RR^d$ such that $F=f$ on $D$ and 
$$\lVert f\rVert_{W_\infty^r(\TT^d)}\leq C_2(d,r)\lVert F\rVert_{W_\infty^r(D)},$$
where $C_2(d,r)$ is a constant that only depends on $d$ and $r$ and $\TT^d=[-\pi,\pi]^d$.

The periodic function $f$ has a Fourier series expansion
$$f(x)=\sum\limits_{k\in\ZZ^d}\hat f(k)e^{ik\cdot x},\qquad x\in\RR^d,$$
where $\left\{\hat f(k)=(2\pi)^{-d}\int_{\TT^d}f(x)e^{-ik\cdot x}dx: k\in \ZZ^d\right\}$ are the Fourier coefficients of $f$.

Let $N$ be an integer. We use the Jackson operator $J_{N,r}$ to approximate $f$.

We first introduce the univariate Jackson kernel as
$$K_{N,r}^{[1]}(t)=\lambda_{N,r}\left(\frac{\sin Mt/2}{\sin t/2}\right)^{2r},\qquad t\in\RR,$$
where $M:=\lfloor N/r\rfloor+1$ with $\lfloor u\rfloor$ being the integer part of $u>0$ and $\lambda_{N,r}=\left[\int_{\TT}(\frac{\sin Mt/2}{\sin t/2})^{2r}dt\right]^{-1}$. The function $K_{N,r}^{[1]}$ is real-valued and $2\pi$-periodic. It also has an expression
$$K_{N,r}^{[1]}(t)=\sum\limits_{k\in\ZZ}\tilde{a}_{k,N}^{[1]}e^{ikt}=\tilde{a}_{0,N}^{[1]}+\sum\limits_{k\in\NN}2\tilde{a}_{k,N}^{[1]}\cos kt,$$
where $\{\tilde{a}_{k,N}^{[1]}\}_{k\in\ZZ}$ is a real-valued even sequence supported in $\{-N,\dots,N\}$. To see this, using prosthaphaeresis formulae and the standard expressions of the $M$-th Fej$\acute{\mathrm{e}}$r kernel
$$\frac{1}{2M}\left(\frac{\sin Mt/2}{\sin t/2}\right)^2=\sum\limits_{\ell=0}^{M-1}b_\ell\cos \ell t,$$
where $b_{j}=1-\frac{j}{M}$ for $j\geq1$ and $b_{0}=\frac{1}{2}$, we can deduce
\begin{equation}
K_{N,r}^{[1]}(t)=\lambda_{N,r}(2M)^{r}\sum\limits_{k=0}^{r(M-1)}\left(\sum\limits_{\epsilon\in\{-1,1\}^{r}}\sum\limits_{\substack{0\leq \ell_i\leq M-1,\ \forall i\\ \lvert\sum\limits_{i=1}^{r}\epsilon_i\ell_i\rvert=k}}\prod\limits_{i=1}^{r}b_{\ell_i}\right)\frac{\cos kt}{2^{r}}.
\end{equation}

The asymptotic behavior $\lambda_{n,r}\sim n^{-2r+1}$ can be found in the literature (e.g., \cite[Chapter 7, Lemma 2.1]{DeVoreLorentz}) and seen easily from the identity $\int_{\TT}(\frac{\sin Mt/2}{\sin t/2})^{2r}dt = 2 \int_{0}^{\pi}(\frac{\sin Mt/2}{\sin t/2})^{2r}dt$ and 
$$\int_{0}^{\pi}(\frac{\sin Mt/2}{t/2})^{2r}dt \leq \int_{0}^{\pi}(\frac{\sin Mt/2}{\sin t/2})^{2r}dt \leq  \int_{0}^{\pi}(\frac{\sin Mt/2}{(2/\pi)\cdot(t/2)})^{2r}dt$$
by bounding the integral $\int_{0}^{\pi}(\frac{\sin Mt/2}{t})^{2r}dt = \left(\frac{2}{M}\right)^{1-2r} \int_{0}^{M\pi/2}(\frac{\sin u}{u})^{2r} du$ as 
$$ \int_{\pi/6}^{5\pi/6}(\frac{\sin u}{u})^{2r} du \leq \int_{0}^{M\pi/2}(\frac{\sin u}{u})^{2r} du \leq \sum_{k=0}^\infty \int_{0}^{2\pi}\frac{(\sin u)^{2r}}{(u+2k\pi)^{2r}}du \leq 2\pi + \sum_{k=1}^\infty \frac{1}{k^{2r}}. $$
Thus, we can bound $\tilde{a}_{k,N}^{[1]}$ as
$$\lvert \tilde{a}_{k,N}^{[1]}\rvert\leq2^{r}\sup\limits_{n\in\NN}(\lambda_{n,r}n^{2r-1}):=C_3(r), \qquad \forall k\in\ZZ.$$

The classical Jackson Theorem (e.g., (2.8) and (2.11) in \cite[Chapter 7]{DeVoreLorentz}) asserts that for the univariate Jackson operator $J_{N,r}^{[1]}$ on $L_\infty(\TT)$ given by
\begin{equation}\label{uniJackson}
J_{N,r}^{[1]}(g,x):=\int_\TT\left[\sum\limits_{\ell=1}^{r}(-1)^{\ell-1}\binom{r}{\ell}g(x+\ell y)\right]K_{N,r}^{[1]}(y)dy,
\end{equation}
there exists a constant $C_4(r)$ depending only on $r$ such that
\begin{equation}\label{uniJackappr}
\lVert J_{N,r}^{[1]}(g)-g\rVert_{L_\infty(\TT)}\leq C_4(r)\|g^{(r)}\|_{L_\infty (\TT)}N^{-r},\qquad\forall g\in W_\infty^r(\TT),\ N\in\NN. 
\end{equation}

Since $\|K_{N,r}^{[1]}\|_{L_1(\TT)}=1$, \eqref{uniJackson} implies
\begin{equation}\label{unimodulusJf}
\|J_{N,r}^{[1]}(g)\|_{L_\infty(\TT)}\leq\sum\limits_{\ell=1}^{r}\binom{r}{\ell}\|g\|_{L_\infty(\TT)}\|K_{N,r}^{[1]}\|_{L_1(\TT)}\leq2^r\|g\|_{L_\infty(\TT)},\quad \forall g\in L_\infty(\TT).
\end{equation}

Observe from a change of variable and the $2\pi$-periodicity of $g$ that 
\begin{equation*}
\begin{split}
J_{N,r}^{[1]}(g,x)=&\sum\limits_{\ell=1}^{r}(-1)^{\ell-1}\binom{r}{\ell} \frac{1}{\ell}\int_0^{2\ell \pi}g(x+t) \sum\limits_{k\in\ZZ}\tilde a_{k,N}^{[1]}e^{ikt/\ell}dt\\
=&\sum\limits_{\ell=1}^{r}(-1)^{\ell-1}\binom{r}{\ell} \frac{1}{\ell}\int_0^{2\pi}g(x+t) \sum\limits_{k\in\ZZ}\tilde a_{k,N}^{[1]}  \sum\limits_{\alpha=0}^{\ell-1} e^{ik(t+2\alpha\pi)/\ell}dt.
\end{split}
\end{equation*}
Notice that the summation $\sum\limits_{\alpha=0}^{\ell-1} e^{ik(t+2\alpha\pi)/\ell}=e^{ikt/\ell} \sum\limits_{\alpha=0}^{\ell-1} \left(e^{i2k\pi/\ell}\right)^\alpha$ vanishes 
when $k\not\in \ell\ZZ$ and equals $\ell e^{ikt/\ell}$ otherwise. Henece 
\begin{equation*}
\begin{split}
J_{N,r}^{[1]}(g,x)
=&\sum\limits_{\ell=1}^{r}(-1)^{\ell-1}\binom{r}{\ell} \int_0^{2\pi}g(x+t) \sum\limits_{k' \in\ZZ}\tilde a_{k' \ell,N}^{[1]}   e^{ik' t}dt \\
=&\int_\TT g(x-y) \sum\limits_{k' \in\ZZ}\sum\limits_{\ell=1}^{r}(-1)^{\ell-1}\binom{r}{\ell} \tilde a_{-k' \ell,N}^{[1]}   e^{ik' y}dy.
\end{split}
\end{equation*}
Thus, by introducing a $2\pi$-periodic kernel $G_{N,r}^{[1]}(t)=\sum\limits_{k \in\ZZ} a_{k,N}^{[1]}e^{ikt}$ with an even sequence of coefficients 
$$a_{k,N}^{[1]}=\sum\limits_{\ell=1}^{r}(-1)^{\ell-1}\binom{r}{\ell}\tilde a_{-k\ell,N}^{[1]},\qquad  k\in\ZZ,$$
we see that the Jackson operator can be expressed as 
$$ J_{N,r}^{[1]}(g,x)=\int_\TT G_{N,r}^{[1]}(y)g(x-y)dy = \int_\TT G_{N,r}^{[1]}(x-y)g(y)dy.$$
The coefficients of the kernel $G_{N,r}^{[1]}$ can be bounded as
\begin{equation}\label{boundcoefficients}
\lvert a_{k,N}^{[1]}\rvert\leq 2^rC_3(r).
\end{equation}

Now we can define a multidimensional $2\pi$-periodic kernel $G_{N,r}$ on $\RR^d$ by
\begin{equation}\label{Jacksonkernel}
G_{N,r}(x)=\prod\limits_{j=1}^dG_{N,r}^{[1]}(x_j)=\sum\limits_{k\in\ZZ^d}a_{k,N}e^{ik\cdot x}.
\end{equation}
where $a_{k,N}=\prod\limits_{j=1}^da_{k_j,N}^{[1]}$ for $k=(k_1,\dots,k_d)\in\ZZ^d$. 
By means of this kernel, we define the Jackson operator on $L_\infty(\TT^d)$ by
\begin{equation}\label{Jnfexplicit}
J_{N,r}(g,x):=\int_{\TT^d}G_{N,r} (x-y) g(y)dy=\sum\limits_{\lVert k\rVert_\infty\leq N}(2\pi)^da_{k,N}\hat g(k)e^{ik\cdot x},\qquad x\in\TT^d.
\end{equation}

To approximate the function $f$, we write $J_{N,r}(f)$ in terms of the multidimensional kernel (\ref{Jacksonkernel}) as $J_{N,r}(f,x) = \int_{\TT^d}\prod\limits_{\ell=1}^{d} G_{N,r}^{[1]}(x_\ell - y_\ell) f(y_1, \ldots, y_d)dy_1 \ldots dy_d$. It is a special case with $j=d$ of the intermediate functions 
$$\left\{\int_{\TT^{j}} \prod\limits_{\ell=1}^{j} G_{N,r}^{[1]}(x_\ell - y_\ell) f(y_1, \ldots, y_{j-1}, y_j, x_{j+1}, \ldots, x_d)dy_1 \ldots dy_{j}\right\}_{j=0}^d$$ while $f(x) = f(x_1, \ldots, x_d)$ corresponds to the case with $j=0$. Then by subtracting and adding the intermediate functions with $j=d-1, \ldots, 1$, we find that the error of 
approximation $J_{N,r}(f) - f$ can be expressed as
$$ J_{N,r}(f,x) - f(x) = \sum_{j=1}^d \int_{\TT^{j-1}} \prod\limits_{\ell=1}^{j-1} G_{N,r}^{[1]}(x_\ell - y_\ell) I_{x_j, \ldots, x_d} (y_1, \ldots, y_{j-1}) dy_1 \ldots dy_{j-1} $$
where $I_{x_j, \ldots, x_d} (y_1, \ldots, y_{j-1})$ is a function on $\TT^{j-1}$ indexed by $x_j, \ldots, x_d \in\RR$ given by 
$$ \int_{\TT} G_{N,r}^{[1]}(x_j - y_j) f(y_1, \ldots, y_{j-1}, y_j, x_{j+1}, \ldots, x_d) dy_j - f(y_1, \ldots, y_{j-1}, x_j, x_{j+1}, \ldots, x_d). $$
Applying (\ref{unimodulusJf}) to the function $\int_{\TT^{j-2}} \prod\limits_{\ell=2}^{j-1} G_{N,r}^{[1]}(x_\ell - y_\ell) I_{x_j, \ldots, x_d} (y_1, \ldots, y_{j-1}) dy_2 \ldots dy_{j-1}$ of the single variable $y_1$, we see that 
\begin{eqnarray*}
&&\sup_{x_1 \in \TT} \left|\int_{\TT^{j-1}} \prod\limits_{\ell=1}^{j-1} G_{N,r}^{[1]}(x_\ell - y_\ell) I_{x_j, \ldots, x_d} (y_1, \ldots, y_{j-1}) dy_1 \ldots dy_{j-1}\right| \\
&\leq& 2^r \sup_{y_1 \in \TT} \left|\int_{\TT^{j-2}} \prod\limits_{\ell=2}^{j-1} G_{N,r}^{[1]}(x_\ell - y_\ell) I_{x_j, \ldots, x_d} (y_1, \ldots, y_{j-1}) dy_2 \ldots dy_{j-1}\right|. 
\end{eqnarray*}
We have by iteration 
\begin{eqnarray*}
&&\sup_{x_1, \ldots, x_{j-1} \in \TT} \left|\int_{\TT^{j-1}} \prod\limits_{\ell=1}^{j-1} G_{N,r}^{[1]}(x_\ell - y_\ell) I_{x_j, \ldots, x_d} (y_1, \ldots, y_{j-1}) dy_1 \ldots dy_{j-1}\right| \\
&\leq& 2^{r(j-1)} \sup_{y_1, \ldots, y_{j-1} \in \TT} \left|I_{x_j, \ldots, x_d} (y_1, \ldots, y_{j-1}) \right|. 
\end{eqnarray*}
But $I_{x_j, \ldots, x_d} (y_1, \ldots, y_{j-1}) = J_{N,r}^{[1]}(h, x_j) - h(x_j)$ where $h$ is the univariate function $h =f(y_1, \ldots, y_{j-1}, \cdot, x_{j+1}, \ldots, x_d)$ indexed by $y_1, \ldots, y_{j-1}, x_{j+1}, \ldots, x_d$.  Hence, by (\ref{uniJackappr}), 
$$ \left|I_{x_j, \ldots, x_d} (y_1, \ldots, y_{j-1}) \right| \leq 
C_4(r)\|h^{(r)}\|_{L_\infty (\TT)} N^{-r}. $$
Observe that $\|h^{(r)}\|_{L_\infty (\TT)} \leq \lVert f\rVert_{W_\infty^r(\TT^d)}$. Therefore, the bound $C_4(r)\lVert f\rVert_{W_\infty^r(\TT^d)} N^{-r}$ for $\left|I_{x_j, \ldots, x_d} (y_1, \ldots, y_{j-1}) \right|$ is independent of $y_1, \ldots, y_{j-1}, x_j, \ldots, x_d$, and we obtain 
\begin{equation}\label{estimation1}
\left\|J_{N,r}(f)-f\right\|_{L_\infty (\TT^d)} \leq \sum_{j=1}^d 2^{r(j-1)}C_4(r)\lVert f\rVert_{W_\infty^r(\TT^d)} N^{-r} \leq d 2^{r d}C_4(r)\lVert f\rVert_{W_\infty^r(\TT^d)} N^{-r}. 
\end{equation}

Shallow ReLU neural networks can approximate the function $J_{N,r}(f)$ well with error bounds stated in terms of its Fourier coefficients. 

\begin{lemma}\label{BarronFourierseries}
For $k\in\ZZ^d$, let $\widehat{J_N}(k)$ be the Fourier coefficient of $J_{N,r}(f)$ at $k$ satisfying
$$J_{N,r}(f,x)=\sum\limits_{k\in\mathbb Z^d}\widehat{J_N}(k)e^{ik\cdot x}.$$
Then for each $m\in\NN$, there exists a function $f_m(x)=\sum\limits_{k=1}^m\beta_k\sigma(\alpha_k\cdot x-b_k)\in H_m$ such that
\begin{equation}\label{estimation2}
\lVert J_{N,r}(f)-f_m\rVert_{L_\infty(D)}\leq C_5v_{J_N,2}\max\left\{\sqrt{\log m},\sqrt d\right\}m^{-\frac{1}{2}-\frac{1}{d}},
\end{equation}
where $C_5$ is an absolute constant,
\begin{equation}
v_{J_N,2}:=\sum\limits_{k\in\mathbb Z^d}\lvert\widehat{J_N}(k)\rvert\lVert k\rVert_1^2,
\end{equation}
and $\beta_k, b_k \in\RR$, $\alpha_k \in\RR^d$ can be bounded as 
$$\lvert\beta_k\rvert\leq\frac{8\pi^2v_{J_N,2}}{m},\qquad\lVert\alpha_k\rVert_1\leq1,\qquad 0\leq b_k\leq1,\qquad\forall \ k=1, \ldots, m.$$
\end{lemma}
The proof of Lemma \ref{BarronFourierseries} is similar to that in \cite{KlusowskiBarron} and is given in details in the appendix.

The Jackson operator used in this paper may be replaced by some other approximation schemes of the form $I_N (f) =\int_{\TT^d} G_N (\cdot, y) f(y) d y$, where $G_N$ is a family of kernels with a scaling index $N$. What is challenging is to approximate $I_N (f)$ by the output $f_m$ of a shallow network induced by ReLU and to estimate the error. This is realized in our approach by a key identity (\ref{eizidentity}) for the function $e^{i z}$ valid in the range $\lvert z\rvert\leq c$ and a concentration inequality for suprema of empirical processes followed by a novel bound for the quantity $v_{J_N,2}$ given in the next section. Another possible approach \cite{Mhaskar2020} is to use some kernels defined by formulae (6.1), (6.26) in \cite{Chandrasekaran} and then apply the related estimates given in Lemma 6.1 and Proposition 4.1 there.  It would be interesting to use such an approach and derive rates of approximating $F\in W^r_\infty(D)$ with a non-integer index $r>0$, which extends our result in Theorem \ref{mainresult}.

\section{Proof of the Main Results by Fourier Analysis}\label{proof of main results}

The key analysis of this paper concentrates on estimating the quantity $v_{J_N,2}$.

Recall $\widehat{J_N}(k)= (2\pi)^da_{k, N} \hat f(k)$ is nonzero only when $\lVert k\rVert_\infty\leq N$ by (\ref{Jnfexplicit}). Let $L=\lceil\log_2N\rceil$, where $\lceil u\rceil$ denotes the smallest integer no less than $u>0$. Then $N\leq2^L\leq2N$. Applying (\ref{boundcoefficients}) to $a_{k,N}=\prod\limits_{j=1}^da_{k_j,N}^{[1]}$ and noticing the term with $k=0$ vanishes, we have
\begin{equation*}
v_{J_N,2}\leq(2^{r+1}\pi C_3(r))^d\sum\limits_{k\in\ZZ^d}\lvert  \hat f(k)\rvert\lVert k\rVert_1^2\leq (2^{r+1}\pi C_3(r))^d\sum\limits_{\ell=0}^L\sum\limits_{2^{\ell-1}<\lVert k\rVert_\infty\leq2^\ell}\lvert  \hat f(k)\rvert\lVert k\rVert_1^2.
\end{equation*}

If $r\geq2$, by $\lVert k\rVert_1\geq\lVert k\rVert_\infty$, we have $\lVert k\rVert_1^{2-r}\leq\lVert k\rVert_\infty^{2-r}\leq(2^{\ell-1})^{2-r}$ when $2^{\ell-1}<\lVert k\rVert_\infty\leq2^\ell$. If $r=1$, we also have $\lVert k\rVert_1^{2-r}\leq d\lVert k\rVert_\infty^{2-r}\leq d(2^{\ell})=2d(2^{\ell-1})^{2-r}$ when $2^{\ell-1}<\lVert k\rVert_\infty\leq2^\ell$. Hence, in either case,
\begin{equation}\label{association}
v_{J_N,2}\leq2d(2^{r+1}\pi C_3(r))^d\sum\limits_{\ell=0}^L(2^{\ell-1})^{2-r}\sum\limits_{2^{\ell-1}<\lVert k\rVert_\infty\leq2^\ell}\lvert  \hat f(k)\rvert\lVert k\rVert_1^r.
\end{equation}

Inspired by some methods in harmonic analysis \cite{Fefferman,Trigub, Carleson, GuoX}, we define a collection of new functions on $\TT^d$, which is the novelty of our time-frequency anlaysis and plays a key role in our error decomposition: for $\ell=1,\dots,L$, let
$$T_{\ell}f(x)=\sum\limits_{\lVert k\rVert_\infty\leq2^\ell}  \hat f(k)\lVert k\rVert_1^re^{ik\cdot x},\qquad x\in\TT^d.$$
The problem of bounding $v_{J_N,2}$ is then transformed to that of bounding $\sum\limits_{k\in \ZZ^d}\lvert\widehat{T_{\ell}f}(k)\rvert$, where $\widehat{T_{\ell}f}(k)= \hat f(k)\lVert k\rVert_1^r$ are the Fourier coefficients of $T_{\ell}f$.

By Parseval's identity,
$$(2\pi)^{-d}\int_{\TT^d}\lvert T_\ell f(x)\rvert^2dx=\sum\limits_{k\in\ZZ^d}\lvert\widehat{T_{\ell}f}(k)\rvert^2,$$
we have
\begin{align}\label{summability1}
\sum\limits_{k\in \ZZ^d}\lvert\widehat{T_{\ell}f}(k)\rvert=&\sum\limits_{\lVert k\rVert_\infty\leq2^\ell}\lvert\widehat{T_{\ell}f}(k)\rvert \leq \left(\sum\limits_{\lVert k\rVert_\infty\leq2^\ell}\lvert\widehat{T_{\ell}f}(k)\rvert^2\right)^{1/2} \left(\sum\limits_{\lVert k\rVert_\infty\leq2^\ell}1^2\right)^{1/2} \nonumber\\
\leq&\left(2^{\ell+1}+1\right)^{\frac{d}{2}} \left((2\pi)^{-d}\int_{\TT^d}\lvert T_\ell f(x)\rvert^2dx\right)^{1/2} \leq(2^{\ell+1}+1)^{\frac{d}{2}}\lVert T_\ell f\rVert_\infty.
\end{align}
Thus we only need to estimate $\lVert T_\ell f\rVert_\infty$ to obtain an upper bound for  $v_{J_N,2}$. This is the main analysis in the proof of our main results. It would be interesting to extend our analysis to some other machine learning algorithms which involve spectral decompositions and frequency analysis \cite{GuoZC, HuT, HuangFengWu}.

\begin{proof}[Proof of Theorem \ref{mainresult}]
Our analysis is based on dividing the set $$U_{\ell}:=\left\{-2^{\ell},-2^{\ell}+1,\dots,2^{\ell}-1,2^{\ell}\right\}^d$$ into disjoint subsets according to the signs of its components as
\begin{equation}\label{divideUell}
U_{\ell}=\bigcup\limits_{\epsilon\in\{-1,0,1\}^d}\Xi_\epsilon,
\end{equation}
where for each $\epsilon=(\epsilon_1,\dots,\epsilon_d)\in\{-1,0,1\}^d$,
$$\Xi_\epsilon=\left\{k=(k_1,\dots,k_d)\in U_\ell:\ \hbox{sgn}(k_j)=\epsilon_j,\ \forall j=1, \ldots, d\right\}.$$
Then
$$T_\ell f(x)=\sum\limits_{\epsilon\in\{-1,0,1\}^d}\sum\limits_{k\in\Xi_\epsilon}  \hat f(k)\lVert k\rVert_1^re^{ik\cdot x}.$$
Observe from the multinomial formula that for $k\in\Xi_\epsilon$,
\begin{displaymath}
\begin{split}
\lVert k\rVert_1^r&=\left(\sum\limits_{j=1}^d\lvert k_j\rvert\right)^r=\left(\sum\limits_{j=1}^d\epsilon_jk_j\right)^r\\
&=\sum\limits_{\substack{\alpha_1+\dots+\alpha_d=r\\ \alpha_1,\dots,\alpha_d\in\ZZ_+}}\frac{r!}{\alpha_1!\dots\alpha_d!}\prod\limits_{j=1}^d(\epsilon_jk_j)^{\alpha_j},
\end{split}
\end{displaymath}
where $\epsilon_j^{\alpha_j}$ denotes $1$ when $\epsilon_j=0$, $\alpha_j=0$. So, for each $\epsilon\in\{-1,0,1\}^d$,
\begin{align}\label{analyzesum1}
&\sum\limits_{k\in\Xi_\epsilon}  \hat f(k)\lVert k\rVert_1^re^{ik\cdot x} \nonumber\\
=&\sum\limits_{\substack{\alpha_1+\dots+\alpha_d=r\\ \alpha_1,\dots,\alpha_d\in\ZZ_+}}\frac{r!}{\alpha_1!\dots\alpha_d!} \left(\prod\limits_{j=1}^d\epsilon_j^{\alpha_j}\right) \left(\sum\limits_{k\in\Xi_\epsilon} \left(\prod\limits_{j=1}^dk_j^{\alpha_j}\right)  \hat f(k)e^{ik\cdot x}\right).
\end{align}
Putting $\hat f(k)=(2\pi)^{-d}\int_{\TT^d}f(t)e^{-ik\cdot t}dt$ into the above sum over $\Xi_\epsilon$,
we have
\begin{align}\label{analyzesum2}
&\sum\limits_{k\in\Xi_\epsilon} \left(\prod\limits_{j=1}^dk_j^{\alpha_j}\right)  \hat f(k)e^{ik\cdot x}
=(2\pi)^{-d}\sum\limits_{k\in\Xi_\epsilon} e^{ik\cdot x}\int_{\TT^d} \left(\prod\limits_{j=1}^dk_j^{\alpha_j}\right) f(t)e^{-ik\cdot t}dt\nonumber\\
=&(2\pi)^{-d}\sum\limits_{k\in\Xi_\epsilon} \int_{\TT^d}(-i)^r\frac{\partial^rf}{\partial x_1^{\alpha_1}\dots\partial x_d^{\alpha_d}}(t)e^{ik\cdot(x-t)}dt\nonumber\\
=&(2\pi)^{-d}(-i)^r\int_{\TT^d}\frac{\partial^r f}{\partial x_d^{\alpha_d}\dots\partial x_1^{\alpha_1}}(t)\prod\limits_{j=1}^d\left(\sum\limits_{k_j\in\widetilde\Xi_\epsilon^{[j]}}e^{ik_j(x_j-t_j)}\right)dt,
\end{align}
where $\widetilde\Xi_\epsilon^{[j]}=\left\{\gamma\in\{-2^\ell,\dots,2^\ell\}:\ \hbox{sgn}(\gamma)=\epsilon_j\right\}$ for $j\in\{1,\dots,d\}$.

If $\epsilon_j=1$ or $-1$, then $\sum\limits_{k_j\in\widetilde\Xi_\epsilon^{[j]}}e^{ik_j(x_j-t_j)}$ equals to $\sum\limits_{\beta=1}^{2^\ell}e^{i\epsilon_j\beta(x_j-t_j)}$. Observe that 
\begin{eqnarray*}
\int_\TT\left|\sum\limits_{\beta=1}^{2^\ell}e^{i\beta t}\right|dt &\leq&\int_\TT\left|\sum\limits_{\beta=1}^{2^\ell}\cos\beta t\right|dt+\int_\TT\left|\sum\limits_{\beta=1}^{2^\ell}\sin\beta t\right|dt\\
&=&\int_\TT\left|\sum\limits_{\beta=1}^{2^\ell}\cos\beta t\right|dt+\int_\TT\left|\frac{2\sin\frac{2^{\ell}+1}{2}t}{2\sin\frac{t}{2}}\sin2^{\ell-1}t\right|dt. 
\end{eqnarray*}
Hence 
\begin{eqnarray*}
\int_\TT\left|\sum\limits_{\beta=1}^{2^\ell}e^{i\beta t}\right|dt &\leq& \int_\TT\left|\frac{1}{2}+\sum\limits_{\beta=1}^{2^\ell}\cos\beta t-\frac{1}{2}\right|dt+\int_\TT\left|\frac{\sin\frac{2^{\ell}+1}{2}t}{\sin\frac{t}{2}}\right|dt\\
&\leq&\int_\TT\left|\frac{1}{2}D_{2^\ell}(t)-\frac{1}{2}\right|dt+\int_\TT\left|D_{2^{\ell-1}}(t)\right|dt,
\end{eqnarray*}
where $D_n$ with $n\in\NN$ is the Dirichlet kernel on $\TT$ given by  $D_n(t)=1+2\sum\limits_{k=1}^n\cos kt$. Since the Dirichlet kernel can be bounded \cite{DeVoreLorentz} as $\lVert D_n\rVert_1\leq(\frac{4}{\pi}\log n+2\pi+1)$, we have
$$\int_\TT\left|\sum\limits_{k_j\in\widetilde\Xi_\epsilon^{[j]}}e^{ik_j(x_j-t_j)}\right|dt=\int_\TT\left|\sum\limits_{\beta=1}^{2^\ell}e^{i\beta t}\right|dt\leq\pi+\frac{1}{2}\lVert D_{2^\ell}\rVert_1+\lVert D_{2^{\ell-1}}\rVert_1\leq (4\pi+2)(\ell+1).$$

If $\epsilon_j=0$, the term in (\ref{analyzesum1}) with $\alpha_j>0$ vanishes. When $\alpha_j=0$, by (\ref{boundcoefficients}),
$$\sum\limits_{k_j\in\widetilde\Xi_\epsilon^{[j]}}e^{ik_j(x_j-t_j)}=1.$$
Therefore, we can bound the $L_1$-norm of the product term in (\ref{analyzesum2}) as
$$\int_{\TT^d}\left|\prod\limits_{j=1}^d\left(\sum\limits_{k_j\in\widetilde\Xi_\epsilon^{[j]}}e^{ik_j(x_j-t_j)}\right)\right|dt=\prod\limits_{j=1}^d\int_\TT\left|\sum\limits_{k_j\in\widetilde\Xi_\epsilon^{[j]}}e^{ik_j(x_j-t_j)}\right| dt_j\leq(4\pi+2)^d(\ell+1)^d.$$
Combining this with (\ref{analyzesum1}) and (\ref{analyzesum2}), we obtain
\begin{eqnarray*}
&&\left|\sum\limits_{k\in\Xi_\epsilon}  \hat f(k)\lVert k\rVert_1^re^{ik\cdot x}\right|\leq\sum\limits_{\substack{\alpha_1+\dots+\alpha_d=r\\ \alpha_1,\dots,\alpha_d\in\ZZ_+}}\frac{r!}{\alpha_1!\dots\alpha_d!}\left|\sum\limits_{k\in\Xi_\epsilon}\prod\limits_{j=1}^d\epsilon_j^{\alpha_j}k_j^{\alpha_j}  \hat f(k)e^{ik\cdot x}\right|\\
&\leq&(2\pi)^{-d}\sum\limits_{\substack{\alpha_1+\dots+\alpha_d=r\\ \alpha_1,\dots,\alpha_d\in\ZZ_+}}\frac{r!}{\alpha_1!\dots\alpha_d!}(4\pi+2)^d\left\lVert\frac{\partial^r f}{\partial x_d^{\alpha_d}\dots\partial x_1^{\alpha_1}}\right\rVert_\infty(\ell+1)^d.
\end{eqnarray*}
It follows that 
\begin{align}\label{Tinfty}
\lVert T_\ell f\rVert_\infty\leq C_6(d,r)\lVert f\rVert_{W_\infty^r(\mathbb T^d)}(\ell+1)^d, 
\end{align}
where $C_6(d,r):=3^d(2\pi)^{-d}\sum\limits_{\substack{\alpha_1+\dots+\alpha_d=r\\ \alpha_1,\dots,\alpha_d\in\ZZ_+}}\frac{r!}{\alpha_1!\dots\alpha_d!}(4\pi+2)^d=(\frac{3}{\pi}+6)^dd^r$.

Combining (\ref{association}), (\ref{summability1}), and (\ref{Tinfty}) yields
\begin{eqnarray*}
v_{J_N,2}&\leq&2d(2^{r+1}\pi C_3(r))^d\sum\limits_{\ell=0}^L(2^{\ell-1})^{2-r}\sum\limits_{k\in\ZZ^d}\lvert\widehat{T_{\ell}f}(k)\rvert\\
&\leq&2d(2^{r+1}\pi C_3(r))^dC_6(d,r)\lVert f\rVert_{W_\infty^r(\mathbb T^d)}\sum\limits_{\ell=0}^L(2^{\ell-1})^{2-r}(2^{\ell+1}+1)^{\frac{d}{2}}(\ell+1)^d\\
&\leq&2d (2^{r+1}\pi C_3(r))^dC_6(d,r)\lVert f\rVert_{W_\infty^r(\mathbb T^d)}  2^{r-2}\times3^{\frac{d}{2}} \sum\limits_{\ell=0}^L(\ell+1)^d(2^\ell)^{\frac{d}{2}+2-r}.
\end{eqnarray*}

For $r<\frac{d}{2}+2$, we have $\frac{d}{2}+2-r >0$. But $r\in\NN$. Hence $\frac{d}{2}+2-r\geq\frac{1}{2}$. It follows that 
\begin{displaymath}
\begin{split}
&\sum\limits_{\ell=0}^L(\ell+1)^d(2^\ell)^{\frac{d}{2}+2-r}\leq(L+1)^d\sum\limits_{\ell=0}^L(2^\ell)^{\frac{d}{2}+2-r}\\
=&(L+1)^d\frac{2^{\frac{d}{2}+2-r}}{2^{\frac{d}{2}+2-r}-1}\cdot\frac{\left(2^{\frac{d}{2}+2-r}\right)^{L+1}-1}{2^{\frac{d}{2}+2-r}}\leq(\log_2N+2)^d\frac{\sqrt2}{\sqrt2-1}(2^{\frac{d}{2}+2-r})^L\\
\leq&\frac{\sqrt2}{\sqrt2-1}(\log_2N+2)^d(2N)^{\frac{d}{2}+2-r}.
\end{split}
\end{displaymath}

For $r=\frac{d}{2}+2$,
$$\sum\limits_{\ell=0}^L(\ell+1)^d(2^\ell)^{\frac{d}{2}+2-r}=\sum\limits_{\ell=0}^L(\ell+1)^d\leq (L+1)^{d+1}\leq(\log_2N+2)^{d+1}.$$

For $r>\frac{d}{2}+2$, we estimate the sum  
$$\sum\limits_{\ell=0}^L(\ell+1)^d(2^\ell)^{\frac{d}{2}+2-r} = \sum\limits_{\ell=0}^L \left\{(\ell+1)^d\left[2^{\frac{1}{2}\left(r-\frac{d}{2}-2\right)}\right]^{-\ell} \cdot  \left[2^{\frac{1}{2}\left(r-\frac{d}{2}-2\right)}\right]^{-\ell}\right\}$$
via bounding the factor $(\ell+1)^d\left[2^{\frac{1}{2}\left(r-\frac{d}{2}-2\right)}\right]^{-\ell}$, $\ell \in \{0, 1, \ldots, L\}$, by $\max\limits_{0\leq\ell\leq L}\left\{(\ell+1)^d\left[2^{\frac{1}{2}\left(r-\frac{d}{2}-2\right)}\right]^{-\ell}\right\}$, and the sum of the remaining terms as 
$$\sum\limits_{\ell=0}^L\left[2^{\frac{1}{2}\left(r-\frac{d}{2}-2\right)}\right]^{-\ell}\leq1+\int_0^L\left[2^{\frac{1}{2}\left(r-\frac{d}{2}-2\right)}\right]^{-t}dt\leq1+\frac{2}{(r-2-\frac{d}{2})\log2}.$$
To estimate the above maximum value, we consider a function $h: (-1, \infty)\to\RR$ given by 
$$ h(t) = (t+1)^d\left[2^{\frac{1}{2}\left(r-\frac{d}{2}-2\right)}\right]^{-t}. $$
From the derivative $h'(t) = (t+1)^{d-1} \left[2^{\frac{1}{2}\left(r-\frac{d}{2}-2\right)}\right]^{-t} \left\{d - (t+1)\log 2^{\frac{1}{2}\left(r-\frac{d}{2}-2\right)}\right\}$, we see that $h$ increases on $(-1, t^{*})$ with $t^{*} =-1 + d/\left(\frac{1}{2}\left(r-\frac{d}{2}-2\right)\log 2\right)$, achieves its maximum value at $t^{*}$, and then decreases on $(t^{*}, \infty)$. Hence 
$$ \max\limits_{0\leq\ell\leq L}\left\{(\ell+1)^d\left[2^{\frac{1}{2}\left(r-\frac{d}{2}-2\right)}\right]^{-\ell}\right\} \leq h(t^{*}) \leq \left(\frac{2 d}{\left(r-\frac{d}{2}-2\right)\log 2}\right)^d 2^{\frac{1}{2}\left(r-\frac{d}{2}-2\right)}. $$
Therefore,
\begin{displaymath}
\sum\limits_{\ell=0}^L(\ell+1)^d(2^\ell)^{\frac{d}{2}+2-r}
\leq \left(1+\frac{2}{(r-2-\frac{d}{2})\log2}\right) \left(\frac{2 d}{\left(r-\frac{d}{2}-2\right)\log 2}\right)^d 2^{\frac{1}{2}\left(r-\frac{d}{2}-2\right)}.
\end{displaymath}
Together with (\ref{estimation1}) and (\ref{estimation2}), the above estimates yield 
\begin{eqnarray*}
\lVert f-f_m\rVert_{L_\infty(D)}&\leq&\lVert f-J_{N,r}(f)\rVert_{L_\infty(D)}+\lVert J_{N,r}(f)-f_m\rVert_{L_\infty(D)}\\
&\leq&d2^{rd}C_4(r)\lVert f\rVert_{W_\infty^r(\TT^d)}N^{-r} +\frac{2\sqrt2}{\sqrt2-1}d 6^{\frac{d}{2}}(2^{r+1}\pi C_3(r))^dC_5C_6(d,r) \\
&\times& \lVert f\rVert_{W_\infty^r(\mathbb T^d)}\max\left\{\sqrt d,\sqrt{\log m}\right\}m^{-\frac{1}{2}-\frac{1}{d}}\\
&\times& \left\{\begin{array}{ll}
(\log_2N+2)^dN^{\frac{d}{2}+2-r},\qquad&\hbox{if }r<\frac{d}{2}+2,\\
(\log_2N+2)^{d+1},\qquad&\hbox{if }r=\frac{d}{2}+2,\\
\left(1+\frac{2}{(r-2-\frac{d}{2})\log2}\right) \left(\frac{2 d}{\left(r-\frac{d}{2}-2\right)\log 2}\right)^d 2^{\frac{3}{2}\left(r-\frac{d}{2}-2\right)},&\hbox{if }r>\frac{d}{2}+2.
\end{array}\right.
\end{eqnarray*}

Finally, by choosing $N=\lfloor m^{\frac{1}{d}(\frac{d+2}{\max\{2r,d+4\}})}\rfloor$ and noting $\lVert f\rVert_{W_\infty^r(\TT^d)}\leq C_2(d,r)\lVert F\rVert_{W_\infty^r(D)}$, we have 
$$\lVert f-f_m\rVert_{L_\infty(D)}\leq C(d,r)\lVert F\rVert_{W_\infty^r(D)}\left\{\begin{array}{ll}
(\log m)^{\frac{1}{2}+d}m^{-\frac{r}{d}\frac{d+2}{d+4}},\qquad&\hbox{if }r<\frac{d}{2}+2,\\
(\log m)^{\frac{3}{2}+d}m^{-\frac{r}{d}\frac{d+2}{d+4}},\qquad&\hbox{if }r=\frac{d}{2}+2,\\
(\log m)^{\frac{1}{2}}m^{-\frac{1}{2}-\frac{1}{d}},\qquad&\hbox{if }r>\frac{d}{2}+2,\end{array}\right.
$$
where
\begin{displaymath}
\begin{split}
C(d,r)=& 2 dC_2(d,r)2^{rd}C_4(r) + \frac{2\sqrt2 d}{\sqrt2-1} 6^{\frac{d}{2}}C_2(d,r)(2^{r+1}\pi C_3(r))^dC_5C_6(d,r)\sqrt d\\
&\times\left\{\begin{array}{ll}
1,&\hbox{if }r\leq\frac{d}{2}+2,\\
\left(1+\frac{2}{(r-2-\frac{d}{2})\log2}\right) \left(\frac{2 d}{\left(r-\frac{d}{2}-2\right)\log 2}\right)^d 2^{\frac{3}{2}\left(r-\frac{d}{2}-2\right)},&\hbox{if }r>\frac{d}{2}+2. \end{array}\right.
\end{split}
\end{displaymath}
Since $f=F$ on $D$, this verifies the desired estimate (\ref{maininequation}) and completes the proof of the theorem.
\end{proof}

As pointed out by a referee, another way to bound $T_\ell f(x)$ is to 
view $e^{ik\cdot x}$ as a univiate function 
of the variable $k\cdot x$ and express it using a $2\pi$ periodic function 
which equals the hat function on $[-1, 1]$ and vanishes on $[-\pi, \pi]\setminus [-1, 1]$. Then some probability estimates might be used to carry out the analysis. 

We can now apply Theorem \ref{mainresult} and the construction in \cite{Zhou, ZhouAA, ZhouDownsampling} to prove our rates of approximation by deep CNNs.

\begin{proof}[Proof of Corollary \ref{dcnncorollary}]
Let $J\geq\frac{2d}{s-1}$ and $m=\lfloor\frac{(s-1)J}{d}-1\rfloor$. By Theorem \ref{mainresult}, we have $f_m \in H_m$ 
with $f_m(x)=\sum\limits_{i=1}^m\beta_i\sigma(\alpha_i\cdot x-t_i)$ on $D$ such that 
$$\lVert F-f_m\rVert_{L_\infty(D)}\leq  \left\{\begin{array}{ll}
C(d,r)\lVert F\rVert_{W_\infty^r(D)}(\log m)^{\frac{1}{2}+d}m^{-\frac{r}{d}\frac{d+2}{d+4}},\qquad&\hbox{if }r<\frac{d}{2}+2,\\
C(d,r)\lVert F\rVert_{W_\infty^r(D)}(\log m)^{\frac{3}{2}+d}m^{-\frac{r}{d}\frac{d+2}{d+4}},\qquad&\hbox{if }r=\frac{d}{2}+2,\\
C(d,r)\lVert F\rVert_{W_\infty^r(D)}(\log m)^{\frac{1}{2}}m^{-\frac{1}{2}-\frac{1}{d}},\qquad&\hbox{if }r>\frac{d}{2}+2.\end{array}\right.$$
Now we realize $f_m$ by an output function $f_J^{\mathbf{w,b}}$ of a deep CNN of depth $J$ constructed 
in \cite[Proof of Theorem 2]{Zhou}. Precisely, first applying \cite[Theorem 3]{Zhou} to the sequence $W=(W_k)_{-\infty}^\infty$ supported in $\{0,\dots,md-1\}$ with
$$[W_{md-1}\ \dots\ W_1\ W_0]=[\alpha_m^\top\ \dots\ \alpha_2^\top\ \alpha_1^\top],$$
adding delta filter sequences at the end if necessary, we can obtain filters $\mathbf w=\{w^{(j)}\}_{j=1}^{J}$ supported in $\{0,\dots,s\}$ such that $W=w^{(J)}*w^{(J-1)}*\dots*w^{(2)}*w^{(1)}$.

Next, taking $\mathbf b=\{b^{(j)}\}_{j=1}^{J-1}$ such that for $j\in\{1,\dots,J-1\}$ and $x\in D$, the components of $T^{w^{(j)}}h^{(j-1)}(x)-b^{(j)}$ are positive.

Finally, for $k=1,\ldots, m$, let $b^{(J)}_{kd}$ be the number which makes the constant term of $\left(h^{(J)}(x)\right)_{kd}$ equals to $t_k$. Taking $c=\left(\sum\limits_{k=1}^m\beta_k \delta_{kd}(j)\right)_{j=1}^{d_J}$ with $\delta_i$ being the delta sequence at $i$ yields $f^{\mathbf{w,b}}_J$. Then we have $f^{\mathbf{w,b}}_J=f_m$.
Combining this identity with the fact that $\frac{1}{2}(s-1)J\leq md\leq(s-1)J$ gives
$$\lVert F-f_J^{\mathbf{w,b}}\rVert_{C(\Omega)}\leq\left\{\begin{array}{ll}
 C_1(d,r)\lVert F\rVert_{W_\infty^r(D)}(\log J)^{d+1/2}J^{-\frac{r}{d}\frac{d+2}{d+4}},\qquad&\hbox{if }r<\frac{d}{2}+2,\\
 C_1(d,r)\lVert F\rVert_{W_\infty^r(D)}(\log J)^{d+3/2}J^{-\frac{r}{d}\frac{d+2}{d+4}},\qquad&\hbox{if }r=\frac{d}{2}+2,\\
C_1(d,r)\lVert F\rVert_{W_\infty^r(D)}(\log m)^{\frac{1}{2}}m^{-\frac{1}{2}-\frac{1}{d}},\qquad&\hbox{if }r>\frac{d}{2}+2,\end{array}\right.$$
where $C_1(d,r)=C(d,r)(2d)^{\frac{r}{d}\frac{d+2}{d+4}}$. This proves Corollary \ref{dcnncorollary}. 
\end{proof}

\section{Discussion}\label{discussion} 

The rate of uniformly approximating functions $f \in W_\infty^r([-1,1]^d))$ given in Theorem \ref{mainresult} is very close to the following lower bound \cite[Theorem 4.2]{DeVoreHowardMicchelli} for neural networks when the data dimension $d$ is large: 

\emph{Let $d, r \in\NN$. Then there exists a constant $c_r$ depending only on $r$ such that for any $n\in\NN$, map $\eta:\mathbb R^n\to C(D)$, and continuous map $M: W_\infty^r([-1, 1]^d)\to\mathbb R^n$ there holds 
$$\sup\limits_{\lVert f\rVert_{W_\infty^r([-1, 1]^d)}\leq1}\lVert f-\eta(M(f))\rVert_\infty\geq c_r n^{-r/d}.$$
}

We apply this lower bound to our setting. If we denote $H_m^p$ with $m\in\NN$ to be the set of output functions $f_m$ on $D$ 
constructed by ReLU deep neural networks of depth $p$ with $(d+1)m$ free parameters, then $H_m^1=H_m$. Let $A_m$ be the collection of these parameters. If there exists a continuous map $M: W_\infty^r(D) \mapsto A_m$, then for any map $\eta: A_m \to C(D)$ which together with $M$ produces $f_m = \eta(M(F))$ there holds
\begin{equation}\label{lowerbound}
\sup\limits_{\lVert F\rVert_{W_\infty^r(D)}\leq1}\lVert F-f_m\rVert_{L_\infty(D)}\geq c_r m^{-\frac{r}{d}}.
\end{equation}

We end our discussion by remarking that when $r>d/2+2$, our main result correponds to that in \cite{KlusowskiBarron}. This was used in \cite{Zhou}. To make it explicit, let
$$W_2^r(\RR^d):=\left\{f\in L_2(\RR^d):\ \lVert f\rVert_{W_2^r(\RR^d)}:=\int_{\RR^d}\lvert\hat f(\omega)\rvert^2(1+|\omega|^{2r})d\omega<\infty\right\},$$
where $\hat f(\omega)=(2\pi)^{-d}\int_{\RR^d}f(x)e^{-i\omega \cdot x}d\omega$ is the Fourier transform of $f$.

By the extension theorem, there exists a constant $C_7 (d,r)$ depending only on $d$ and $r$ such that each $F\in W_\infty^r(D)$ can be extended to a function $f$ in $W^r_2(\RR^d)$ with
$$\lVert f\rVert_{W_2^r(\RR^d)}\leq C_7 (d,r)\lVert F\rVert_{W_2^r(D)}\leq(2\pi)^{d/2}C_7 (d,r)\lVert F\rVert_{W_\infty^r(D)}.$$
Then $r>d/2+2$ is a sufficient condition for the finiteness of $v_{f, 2}$ defined in \cite{KlusowskiBarron} as 
$$v_{f,2}:=\int_{\RR^d}\lvert\hat f(\omega)\rvert\lVert\omega\rVert_1^2d\omega<\infty. $$
In fact, we have 
\begin{displaymath}
\begin{split}
v_{f,2}=&\int_D\lvert\hat f(\omega)\rvert\lVert\omega\rVert_1^2d\omega+\int_{\RR^d\setminus D}\lvert\hat f(\omega)\rvert\lVert\omega\rVert_1^2d\omega\\
\leq&\left(\int_Dd^2\lvert\hat f(\omega)\rvert^2 \left|\omega\right|^4 d\omega\cdot\int_D1d\omega\right)^{1/2} \\
&+\left(\int_{\RR^d\setminus D}d^{2r}\lvert\hat f(\omega)\rvert^2\left|\omega\right|^{2r}d\omega\cdot\int_{\RR^d\setminus D}\lVert\omega\rVert_1^{4-2r}d\omega\right)^{1/2}\\
\leq&C_8 (d,r)\left(\int_{\RR^d}\lvert\hat f(\omega)\rvert^2 \left(1+\left|\omega\right|^{2r}\right) d\omega\right)^{1/2}\leq(2\pi)^{d/2}C_7 (d,r)C_8 (d,r)\lVert F\rVert_{W_\infty^r(D)},
\end{split}
\end{displaymath}
where $C_8 (d,r)$ is a constant depending only on $d$ and $r$. Then for $f\in W_\infty^r(D)$,
\begin{equation}\label{comparison}
\inf\limits_{f_m\in H_m}\lVert f-f_m\rVert_{L_\infty(D)}\leq C_{9} (d,r)\lVert f\rVert_{W_\infty^r(D)}(\log m)^{\frac{1}{2}}m^{-\frac{1}{2}-\frac{1}{d}},
\end{equation}
where $C_{9}(d,r)$ only depends on $d$ and $r$. This is the rate of approximation we obtained in (\ref{maininequation}).

When $r=d/2+2$, we have $-\frac{r}{d}\frac{d+2}{d+4}=-\frac{1}{2}-\frac{1}{d}$. Then the upper bound in Theorem \ref{mainresult} is the same as that in \cite{KlusowskiBarron} up to a logarithmic term. 

\section*{Appendix}

In this appendix, we provide a detailed proof of Lemma \ref{BarronFourierseries} for completeness. The method of the proof is borrowed from \cite{KlusowskiBarron}. Here is an outline of the proof: we first represent the Fourier series basis function $e^{i k\cdot x}$ as an integral of the shifts $\sigma (k\cdot x -u)$ of the ReLU multiplied with $e^{i u}$. Then we express the value at $x\in D$ of the Jackson operator $J_{N,r}(f,x)$ as the expectation of a random variable. Finally, we approximate the expectation by an empirical mean. The key part of the proof is to conduct  
Rademacher analysis for estimating the error between the expectation and the empirical mean, uniformly for $x\in D$, by applying a concentration inequality for suprema of empirical processes to a collection of random variables indexed by the set 
$D\times\{-1,0,1\}$.

\begin{proof}[Proof of Lemma \ref{BarronFourierseries}]
The following identity stated as Equation (19) in \cite{KlusowskiBarron} 
\begin{equation}\label{eizidentity}
e^{iz}-iz-1=-\int_0^c\sigma(z-u)e^{iu}+\sigma(-z-u)e^{-iu}du, \qquad \lvert z\rvert\leq c
\end{equation}
applied to $c=\lVert\pi k\rVert_1$ and $z=k\cdot x$ with $k\in \ZZ^d\setminus \{0\}$ yields
$$e^{ik\cdot x}=-\int_0^{\lVert\pi k\rVert_1}\sigma(k\cdot x-u)e^{iu}+\sigma(-k\cdot x-u)e^{-iu}du+ik\cdot x+1.$$
Changing the variable $u$ by $t=\frac{u}{\lVert\pi k\rVert_1}$ and using $\frac{1}{\lVert\pi k\rVert_1}\sigma(v)=\sigma(\frac{v}{\lVert\pi k\rVert_1})$ for $v\in\RR$, we have
$$e^{ik\cdot x}=-\lVert\pi k\rVert_1^2\int_0^1\sigma\left(\frac{k}{\lVert\pi k\rVert_1}\cdot x-t\right)e^{i\lVert\pi k\rVert_1t}+\sigma\left(-\frac{k}{\lVert\pi k\rVert_1}\cdot x-t\right)e^{-i\lVert\pi k\rVert_1t}dt+ik\cdot x+1.$$
Putting this expression into the Jackson operator $J_{N,r}(f,x)=\sum\limits_{k\in\ZZ^d}\widehat{J_N}(k)e^{ik\cdot x}$, we find 
\begin{eqnarray*}
J_{N,r}(f,x)&=&\sum\limits_{k\in\mathbb Z^d}\widehat{J_N}(k) \biggl\{-\lVert\pi k\rVert_1^2\int_0^1\sigma\left(\frac{k}{\lVert\pi k\rVert_1}\cdot x-t\right)e^{i\lVert\pi k\rVert_1t} \\
&& + \sigma\left(-\frac{k}{\lVert\pi k\rVert_1}\cdot x-t\right)e^{-i\lVert\pi k\rVert_1t}dt\biggr\}
+\sum\limits_{k\in\mathbb Z^d}\widehat{J_N}(k)(ik\cdot x+1). 
\end{eqnarray*}
Take a phase $b(k)\in(-\pi,\pi]$ of the complex number $\widehat{J_N}(k)$ satisfying 
$$\widehat{J_N}(k)=\lvert\widehat{J_N}(k)\rvert e^{i b(k)}.$$
Then $\widehat{J_N}(k)  e^{\pm i\lVert\pi k\rVert_1t} = |\widehat{J_N}(k)| e^{i\left(\pm \lVert\pi k\rVert_1t +b(k)\right)}$. Notice that $J_{N,r}(f,x)$ is a real number. Then by taking its real part, we have 
\begin{eqnarray*}
J_{N,r}(f,x)&=&-\pi^2\sum\limits_{k\in\ZZ^d\setminus\{0\}}|\widehat{J_N}(k)|\|k\|_1^2\int_0^1\sigma\left(\frac{k}{\lVert\pi k\rVert_1}\cdot x-t\right)\cos (\lVert\pi k\rVert_1t+b(k))dt\\
&&-\pi^2\sum\limits_{k\in\ZZ^d\setminus\{0\}}|\widehat{J_N}(k)|\|k\|_1^2\int_0^1\sigma\left(-\frac{k}{\lVert\pi k\rVert_1}\cdot x-t\right)\cos (-\lVert\pi k\rVert_1t + b(k))dt\\
&&-\left[\sum\limits_{k\in\mathbb Z^d}\mathrm{Im}(\widehat{J_N}(k))k\right]\cdot x+\sum\limits_{k\in\mathbb Z^d}\mathrm{Re}(\widehat{J_N}(k)). 
\end{eqnarray*}

To approximate the function $J_{N,r}(f, x)$ by the output $f_m(x) =\sum\limits_{k=1}^m\beta_k\sigma(\alpha_k\cdot x-b_k)\in H_m$ of a shallow network, we regard $J_{N,r}(f, x)$ as the expectation of a random variable, discretize it, and then estimate the error by a concentration inequality. Here $x\in D$ is used as an index of a collection of random vaiables. 

We first take a probability measure $P$ on $\{-1,1\}\times[0,1]\times \left(\mathbb Z^d\setminus\{0\}\right)$ by setting for $z\in \{-1,1\}, k\in\ZZ^d \setminus\{0\}$ the density as 
$$p(z,t,k)=\frac{\pi^2}{v}\lvert\widehat{J_N}(k)\rvert \ \lVert k\rVert_1^2 \  \lvert\cos(z\lVert\pi k\rVert_1t+b(k))\rvert, \qquad t\in [0, 1],$$
where $v$ is the normalization constant 
$$v=\pi^2\sum\limits_{k\in\ZZ^d}\left[\lVert k\rVert_1^2\lvert\widehat{J_N}(k)\rvert\int_0^1\lvert \cos (\lVert\pi k\rVert_1t+b(k))\rvert+\lvert \cos (-\lVert\pi k\rVert_1t + b(k))\rvert dt\right]\leq2\pi^2v_{J_N,2}.$$

We then define a collection of random variables $\{h_x\}_{x\in D}$ on 
$\{-1,1\}\times[0,1]\times \left(\mathbb Z^d\setminus\{0\}\right)$ 
given by 
\begin{equation}\label{randomvariables}
h_x (z,t,k) =\sigma(z\alpha\cdot x-t)s(zt,k),\qquad z\in \{-1,1\}, t\in [0,1], k\in \mathbb Z^d\setminus\{0\},
\end{equation}
where $\alpha=\alpha_k :=\frac{k}{\lVert\pi k\rVert_1}$ and $s(t,k):=-\mathrm{sgn}(\cos (\lVert\pi k\rVert_1t+b(k)))$. For each $x\in D$, the expected value $\EE_P[h_x] =\int_{\{-1,1\}\times[0,1]\times\left(\mathbb Z^d\setminus\{0\}\right)}h_x (z,t,k) dP(z, t,k)$ of the random variables $h_x$ satisfies 
\begin{displaymath}
\begin{split}
&J_{N,r}(f,x) + \left[\sum\limits_{k\in\mathbb Z^d}\mathrm{Im}(\widehat{J_N}(k))k\right]\cdot x - \sum\limits_{k\in\mathbb Z^d}\mathrm{Re}(\widehat{J_N}(k))\\
=&v\int_{\{-1,1\}\times[0,1]\times\mathbb Z^d\setminus\{0\}}h_x (z,t,k)dP(z, t,k) =:g_x. 
\end{split}
\end{displaymath}
The rest of the proof is analogous to \cite[Proof of Theorem 1]{KlusowskiBarron}, and we replace $m$ by $m'=\lceil m/4\rceil$ here. 

Let $\epsilon >0$ to be determined later. We can partition the set $\{(z,t,\alpha) \in\{-1,1\} \times [0,1] \times \RR^d: \ \|\alpha\|_1=\pi^{-1}\}$ into a family of subsets $\{\mathcal A_j\}_{j=1}^{M'}$ of $\ell^\infty$-diameter at most $\frac{\epsilon}{d+1}$, where the number $M'$ of the subsets in this family can be chosen to be the integer part of  $2\left(\frac{2d+2}{\pi}\right)^{d-1}(d+1)\epsilon^{-d}$. The diamater restriction yields 
$$\sup\limits_{(z,t,\alpha),(\tilde z,\tilde t,\tilde\alpha)\in\mathcal A_j}\|(z,t,\alpha)-(\tilde z,\tilde t,\tilde\alpha)\|_\infty\leq\frac{\epsilon}{d+1},\quad j=1,\dots,M',$$
which together with the Lipschitz property of $\sigma$ implies  
$$\sup\limits_{x\in D}\left|\sigma(z\alpha\cdot x-t)-\sigma(\tilde z\tilde\alpha\cdot x-\tilde t)\right|\leq\epsilon,\quad\forall(z,t,\alpha),(\tilde z,\tilde t,\tilde\alpha)\in\mathcal A_j.$$
For each $j$, we denote two subsets of $\{-1,1\}\times[0,1]\times\left(\ZZ^d\setminus\{0\}\right)$ as
\begin{eqnarray*}
\mathcal A_{j,-}&=&\left\{(z,t,k):\ \left(z,t,\frac{k}{\|\pi k\|_1}\right)\in\mathcal A_j,\ s(zt,k)=-1\right\},\\
\mathcal A_{j,+}&=&\left\{(z,t,k):\ \left(z,t,\frac{k}{\|\pi k\|_1}\right)\in\mathcal A_j,\ s(zt,k)=1\right\},
\end{eqnarray*}
and set the collection $\left\{\mathcal A_{j,-}, \mathcal A_{j,+}: j=1, \ldots, M'\right\}$ as $\{\mathcal B_i\}_{j=1}^M$ with $M=2M'$. Then $\{\mathcal B_1,\dots,\mathcal B_M\}$ form a partition of the set $\Lambda=\{-1,1\}\times[0,1]\times\left(\ZZ^d\setminus\{0\}\right)$ and satisfy 
\begin{equation}\label{partition of Lambda}
\sup\limits_{\left( z, t, k\right),(\tilde z,\tilde t,\tilde k)\in\mathcal B_i} \sup\limits_{x\in D} \left|h_x\left(\tilde z,\tilde t,\tilde k\right)-h_x\left( z,t,k\right)\right| \leq \epsilon,\quad i=1,\dots,M.
\end{equation}
We restrict the probability measure $P$ onto the subsets in this partition and define 
a collection of probability measures $\{P_i\}_{i=1}^M$ on $\Lambda$ by 
$$dP_i (z,t,k)= \frac{1}{L_i} dP(z,t,k)\mathbf1\{(z,t,k)\in\mathcal B_i\}, $$
where $L_i = \int_{\mathcal B_i} dP(z,t,k)$ is the normalization constant to make $P_i$ a probability measure. 
Correspondingly, we set $m_i=m'L_i$ and take a random sample 
$$\underline a=\{(z_{j,i},t_{j,i},k_{j,i})\}_{1\leq j\leq n_i,\ 1\leq i\leq M}$$
of sizes $\{n_i=\lceil m_i\rceil\}$ independently according to $\{P_i\}_{i=1}^M$. Thus, we split the population domain $\Lambda$ into $M$ “strata” $\mathcal B_1,\dots,\mathcal B_M$ and allocate the number of within-stratum samples to be proportional to the “size” of the stratum $m_1,\dots,m_M$ (i.e., proportionate allocation). Note from $\sum\limits_{i=1}^M L_i =1$ that
\begin{equation}\label{sum of nk}
\sum\limits_{i=1}^M n_i\leq m'+M. 
\end{equation}

Let 
$$g_{i, x}=\frac{v}{n_i}\sum\limits_{j=1}^{n_i}h_x (z_{j,i},t_{j,i},k_{j,i}), \qquad i=1,\dots,M$$ 
and 
$$\overline g_{x}=\sum\limits_{i=1}^M \frac{m_i}{m'} g_{i, x}. $$ 
We apply $L_i=m_i/m'$ to get
\begin{eqnarray}\label{divide expectation}
&&\EE\left[\sup\limits_{x\in D}\left|\overline g_{x}-g_x\right|\right]=\EE\left[\sup\limits_{x\in D}\left|\sum\limits_{i=1}^M L_i g_{i, x}-v\sum\limits_{i=1}^ML_i\int_{\mathcal B_i}h_x (z,t,k)dP_i(z,t,k)\right|\right]\nonumber\\
&=&\frac{v}{m'}\EE\left[\sup\limits_{x\in D}\left|\sum\limits_{i=1}^M \frac{m_i}{n_i} \sum\limits_{j=1}^{n_i}\left(h_x (z_{j,i},t_{j,i},k_{j,i})-\EE_{P_i}[h_x]\right)\right|\right].
\end{eqnarray}

To carry out Rademacher analysis for the quantity in (\ref{divide expectation}), we 
let $\underline\sigma=\{\sigma_{j,i}\}$ be a sequence of independent identically distributed Rademacher variables and $\{\mu_{i}\}_{i=1}^M$ be a sequence of functions defined on $D$ by $\mu_i (x)=h_x (z_i,t_i,k_i)$ with a random sample $(z_i,t_i,k_i)\in\mathcal B_i$ drawn according to $P_i$. 
We get from \cite[Lemma 2.3.6]{VaartWellner} that
\begin{eqnarray}\label{expectation for fixed m}
&&\EE\left[\sup\limits_{x\in D}\left|\sum\limits_{i=1}^M \frac{m_i}{n_i} \sum\limits_{j=1}^{n_i}\left(h_x (z_{j,i},t_{j,i},k_{j,i})-\EE_{P_i}[h_x]\right)\right|\right] \nonumber\\
&\leq&2\EE\left[\sup\limits_{x\in D}\left|\sum\limits_{i=1}^M \frac{m_i}{n_i} \sum\limits_{j=1}^{n_i}\sigma_{j,i}\left(h_x (z_{j,i},t_{j,i},k_{j,i})-\mu_i(x)\right)\right|\right].
\end{eqnarray}

For notational brevity, we denote $\tilde h_{j,i}(x)=\frac{m_i}{n_i}(h_x (z_{j,i},t_{j,i},k_{j,i})-\mu_i(x))$. Observe that 
$\sup\limits_{y\in \{-1,0,1\}}\left(\sum\limits_{i=1}^M\sum\limits_{j=1}^{n_i}\sigma_{j,i}y\tilde h_{j,i}(x)\right) = \left|\sum\limits_{i=1}^M\sum\limits_{j=1}^{n_i}\sigma_{j,i}\tilde h_{j,i}(x)\right|$. Fix $\underline a$. We apply a concentration inequality \cite[Corollary 13.2]{BoucheronLugosiMassart} for suprema of empirical processes involving a collection of random variables $\left\{\sum\limits_{k=1}^n \alpha_{k, t} \epsilon_k: t \in \mathcal T\right\}$ induced by a sequence of independent Rademacher variables $\{\epsilon_k\}_{k=1}^n$ and a collection of coefficient sequences $\{\alpha_{k, t}\}_{k=1}^n$ indexed by a set $\mathcal T$ with the distance $\hbox{dist}(t, t') = \left\{\sum_{k=1}^n \left(\alpha_{k, t} - \alpha_{k, t'}\right)^2\right\}^{1/2}$ for $t, t' \in \mathcal T$. In our situation, $\underline\sigma=\{\sigma_{j,i}\}$ is the sequence of independent Rademacher variables.  The collection of coefficient sequences is $\left\{\left(y\tilde h_{j,i}(x)\right)_{j, i}: (x,y)\in \mathcal T\right\}$ indexed by the set $\mathcal T := D\times\{-1,0,1\}$ with the distance $\kappa$ given by  
\begin{eqnarray*}
\kappa \left((x,y),(x',y')\right)=\left(\sum\limits_{i=1}^M\sum\limits_{j=1}^{n_i}\left(y\tilde h_{j,i}(x)-y'\tilde h_{j,i}(x')\right)^2\right)^{1/2}, \qquad (x,y), (x',y')\in {\mathcal T}. 
\end{eqnarray*}
Hence we can apply \cite[Corollary 13.2]{BoucheronLugosiMassart} and obtain 
\begin{eqnarray}\label{expectation on sigma}
&&\EE_{\underline\sigma}\left[\sup\limits_{x\in D}\left|\sum\limits_{i=1}^M\sum\limits_{j=1}^{n_i}\sigma_{j,i}\tilde h_{j,i}(x)\right|\right]\nonumber\\
&=&\EE_{\underline\sigma}\left[\sup\limits_{(x,y)\in D\times\{-1,0,1\}}\left(\sum\limits_{i=1}^M\sum\limits_{j=1}^{n_i}\sigma_{j,i} y\tilde h_{j,i}(x)\right)-0\right]\nonumber\\
&\leq&12\int_0^{\delta/2}\sqrt{N(u, {\mathcal T})}du,
\end{eqnarray}
where $N(u, {\mathcal T})$ is the $u$-metric entropy of ${\mathcal T}$ with respect to the metric $\kappa$ (i.e., the logarithm of the smallest size of $u$-nets that cover ${\mathcal T}$ with respect to $\kappa$) and $\delta=\left(\sup\limits_{(x,y)\in D\times\{-1,0,1\}}\sum\limits_{i=1}^M\sum\limits_{j=1}^{n_i}\left(y\tilde h_{j,i}(x)\right)^2\right)^{1/2}$. 

To estimate the metric entropy $N(u, {\mathcal T})$, we observe from a simple covering of the interval $[-1, 1]$ by $1+1/\eta$ intervals of radius $\eta>0$ that the cube $D=[-1, 1]^d$ can be covered by $(1+1/\eta)^d \leq (2/\eta)^d$ balls of radius $\eta$ in the $\ell_\infty$-norm for $\eta\leq 1$. Combining this with a metric relation 
$$\kappa \left((x,y), (x',y)\right)=|y|\left(\sum\limits_{i=1}^M\sum\limits_{j=1}^{n_i}\left(\tilde h_{j,i}(x)-\tilde h_{j,i}(x')\right)^2\right)^{1/2} \leq 2\sqrt{m'+M}\|x-x'\|_\infty$$
seen from the Lipschitz property of $\sigma$ and the definition of $\tilde h_{j,i}$,  
we find that any $\eta$-covering of $D$ with respect to the $\ell_\infty$-norm induces a $2\sqrt{m'+M}\eta$ covering of $D\times \{y\}$ with respect to the $\kappa$-metric. 
Therefore, by taking $\eta=u/(2\sqrt{m'+M}) \leq \frac{\delta}{2}/(2\sqrt{m'+M}) <1$, for the covering numbers $\mathcal N(u, D\times \{y\})$ and $\mathcal N(u, \mathcal T)$, we have 
$$ 
\mathcal N(u, D\times \{y\})\leq (2/\eta)^d \leq \left(\frac{4 \sqrt{m'+M}}{u}\right)^d $$
and
\begin{equation}\label{kappa norm} \mathcal N(u, \mathcal T)\leq \sum_{y \in \{1, 0, -1\}} \mathcal N(u, D\times \{y\}) \leq 3\left(\frac{4 \sqrt{m'+M}}{u}\right)^d.
\end{equation}
It follows from (\ref{partition of Lambda}) and (\ref{sum of nk}) that $\delta\leq\sqrt{m'+M}\epsilon$ and from (\ref{kappa norm}) that $N(u, \mathcal T)\leq d\log\left(4\sqrt{m'+M}/u\right)+\log3$. 

Now we determine $\epsilon>0$ by 
$$ \epsilon = \frac{2(d+1) \pi^{-1+1/d}}{\lceil m/4\rceil^{1/d}}. $$
This choice together with the definition of $M'$ gives $M' = 2\left((2d+2)/\pi\right)^{d-1} (d+1) \epsilon^{-d} = \lceil m/4\rceil$. Hence  
$M'=m'=\lceil m/4\rceil$ and $M= 2M'=2\lceil m/4\rceil$. Then by evaluating the integral, we can bound (\ref{expectation on sigma}) as 
\begin{eqnarray}\label{first term}
\EE_{\underline\sigma}\left[\sup\limits_{x\in D}\left|\sum\limits_{i=1}^M\sum\limits_{j=1}^{n_i}\sigma_{j,i}\tilde h_{j,i}(x)\right|\right] 
\leq12(3+2\sqrt{\log m})\sqrt d\sqrt{\frac{m}{2}}\epsilon.
\end{eqnarray}
Thus by taking expection over $\underline a \in \Lambda$, we obtain 
\begin{eqnarray}
&&\EE\left[\sup\limits_{x\in D}\left|\sum\limits_{i=1}^M \frac{m_i}{n_i} \sum\limits_{j=1}^{n_i}\sigma_{j,i}\left(h_x (z_{j,i},t_{j,i},k_{j,i})-\mu_i(x)\right)\right|\right]\nonumber\\
&=&\EE_{\underline a}\EE_{\underline\sigma}\left[\sup\limits_{x\in D}\left|\sum\limits_{i=1}^M\sum\limits_{j=1}^{n_i}\sigma_{j,i}\tilde h_{j,i}(x)\right|\right] \leq 12(3+2\sqrt{\log m})\sqrt d\sqrt{\frac{m}{2}}\epsilon.
\end{eqnarray}
Together with (\ref{divide expectation}) and (\ref{expectation for fixed m}), we conclude that 
\begin{equation}
\EE\left[\sup\limits_{x\in D}|\overline g_x-g_x|\right]\leq \frac{C_5}{2\pi^2}vd^{3/2}\sqrt{\log m}m^{-\frac{1}{2}-\frac{1}{d}}
\end{equation}
holds for some absolute constant $C_5$. Since this inequation holds on average, by (\ref{sum of nk}) we know that there is a realization
$$\overline g_x=\sum\limits_{i=1}^M\sum\limits_{j=1}^{n_i}\frac{vm_i}{m'n_i}h_x (z_{j,i},t_{j,i},\omega_{j,i}) =: \sum\limits_{k=1}^{3 \lceil m/4\rceil}  \beta_k\sigma(\alpha_k\cdot x-b_k)  \in H_{3 \lceil m/4\rceil}$$
such that 
\begin{displaymath}
\begin{split}
& \sup\limits_{x\in D}\left| J_{N,r}(f,x)+\left[\sum\limits_{k\in\mathbb Z^d}\mathrm{Im}(\widehat{J_N}(k))k\right]\cdot x-\sum\limits_{k\in\mathbb Z^d}\mathrm{Re}(\widehat{J_N}(k))-\overline g_x\right|\\
\leq &C_5v_{J_N,2}d^{3/2}\sqrt{\log m}m^{-\frac{1}{2}-\frac{1}{d}}. 
\end{split}
\end{displaymath}
Moreover, from the definition (\ref{randomvariables}) of $h_x$, 
we can get bounds of the parameters as 
$$\lvert\beta_k\rvert\leq\frac{v}{\lceil m/4\rceil}\leq \frac{8\pi^2v_{J_N,2}}{m},\qquad\lVert\alpha_k\rVert_1\leq1,\qquad 0\leq b_k\leq1.$$
To complete the proof, notice $u=\sigma(u)-\sigma(-u)$ for $u\in\RR$, then for $m\geq 20$, a function of the form $\sum\limits_{k=3 \lceil m/4\rceil+1}^m\beta_k\sigma(\alpha_k\cdot x-b_k)$ can realize the affine function $\left[\sum\limits_{k\in\mathbb Z^d}\mathrm{Im}(\widehat{J_N}(k))k\right]\cdot x-\sum\limits_{k\in\mathbb Z^d}\mathrm{Re}(\widehat{J_N}(k))$ with the parameters bounded as
$$\lvert\beta_k\rvert\leq\frac{8 v_{J_N,2}}{m},\qquad\lVert\alpha_k\rVert_1\leq1,\qquad 0\leq b_k\leq1,$$
and the desired bound (\ref{estimation2}) is verified. The bound is trivially true for $m<20$. 
This completes the proof of Lemma \ref{BarronFourierseries}. 
\end{proof}

\section*{Acknowledgments} 
The first version of the paper was written when the authors were at City University of Hong Kong, supported partially by 
NSFC/RGC Joint Research Scheme [RGC Project No. N\_CityU102/20 
and NSFC Project No. 12061160462], 
Germany/Hong Kong Joint Research Scheme [Project No. G-CityU101/20], 
Hong Kong Institute for Data Science, 
and InnoHK initiative, The Government of the HKSAR, and Laboratory for AI-Powered Financial Technologies. The authors would like to thank Hrushikesh Mhaskar and the referees for their constructive comments and suggestions.

\bibliographystyle{abbrvnat}

\end{document}